\documentclass[runningheads]{llncs}

\usepackage{graphicx}
\usepackage{amsmath,amssymb} 
\usepackage{color}

\usepackage{times}
\usepackage{epsfig}
\usepackage{bm}
\usepackage{subcaption}
\usepackage{float}

\newcommand{\ignore}[1]{}
\newcommand{\warp}{\omega}

\def\calB{\mathcal{B}}
\def\calC{\mathcal{C}}

\def\calI{\mathcal{I}}

\def\calL{\mathcal{L}}

\def\calN{\mathcal{N}}

\usepackage[pagebackref=true,breaklinks=true,letterpaper=true,colorlinks,bookmarks=false]{hyperref}

\begin{document}
	
	\pagestyle{headings}
	\mainmatter
	\def\ECCV18SubNumber{1517}  
	
	\title{Robust image stitching with multiple registrations}
	
	\titlerunning{Robust image stitching with multiple registrations}
	
	\authorrunning{C. Herrmann, C. Wang, R.S. Bowen, E. Keyder, M. Krainin, C. Liu and R. Zabih}
	
	\author{Charles Herrmann\inst{1} \and Chen Wang\inst{1,2} \and Richard Strong Bowen\inst{1} \and Emil Keyder\inst{2} \and\\ Michael Krainin\inst{3} \and Ce Liu\inst{3} \and Ramin Zabih\inst{1,2}}
	\institute{
		Cornell Tech, New York, NY 10044, USA \and
		Google Research, New York, NY 10011, USA \and
		Google Research, Cambridge, MA 02142, USA\\
		\email{\{cih,chenwang,rsb,rdz\}@cs.cornell.edu, \{wangch,emilkeyder,mkrainin,celiu,raminz\}@google.com}
	}

	\maketitle
	
	\begin{abstract}
		Panorama creation is one of the most widely deployed techniques in computer vision. 
		In addition to industry applications such as Google Street View, it is also used by
		millions of consumers in smartphones and other cameras. Traditionally, the
		problem is decomposed into three phases: {registration}, which picks a single
		transformation of each source image to align it to the other inputs, {seam~finding}, which selects a source image for each pixel in the final result, and blending, which fixes minor visual artifacts \cite{Lin2016,Zhang_2014_CVPR}. Here, we observe that the use of a single registration often leads to errors, especially in scenes with significant depth variation or object motion. We propose instead the use of \textit{multiple} registrations, permitting regions of the image at different depths to be captured with greater accuracy. MRF inference techniques
		naturally extend to seam finding over multiple registrations, and we show here that
		their energy functions can be readily modified with new terms that discourage
		duplication and tearing, common problems that are exacerbated by the use of multiple
		registrations. Our techniques are closely related to layer-based stereo \cite{SergeWhatWentWhere,WangAdelson}, and move image stitching closer to explicit scene modeling.
		Experimental evidence demonstrates that our techniques often generate
		significantly better panoramas when there is substantial motion or parallax.
	\end{abstract}

	\section{Image stitching and parallax errors}
	\label{sec:introduction}
	
\begin{figure}
		\begin{centering}
			\begin{subfigure}[b]{0.8\textwidth}
				\centering
				\includegraphics[height=0.1\textheight]{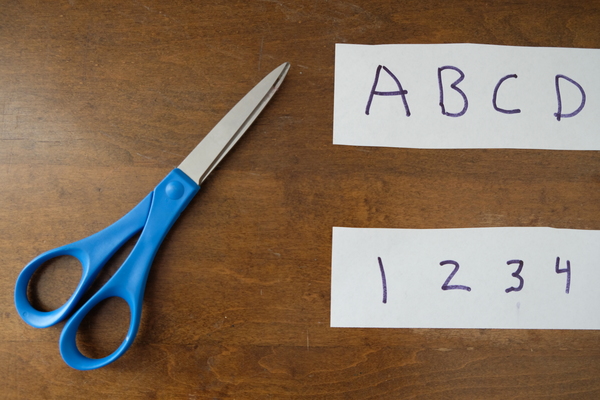}
				\includegraphics[height=0.1\textheight]{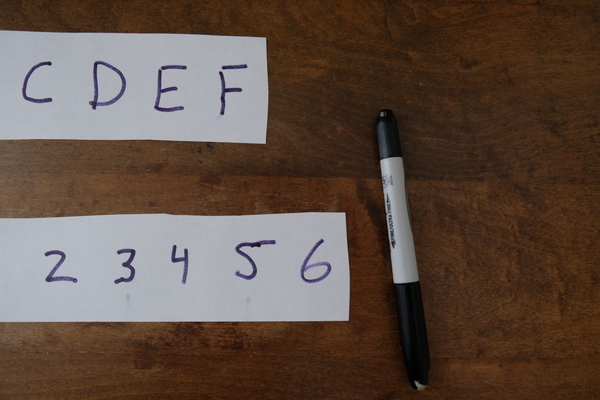}
				\caption{Input images}
				\label{fig:cartoon-inputs}
			\end{subfigure}
			\begin{subfigure}[b]{0.4\textwidth}
				\centering
				\includegraphics[height=0.1\textheight]{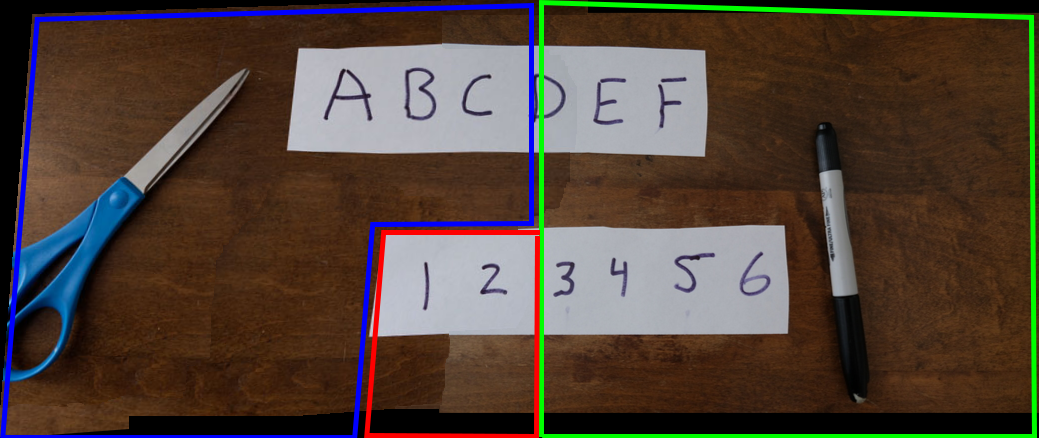}
				\caption{Our result}
				\label{fig:cartoon-align-triangles}
			\end{subfigure}
			\begin{subfigure}[b]{0.4\textwidth}
				\centering
				\includegraphics[height=0.1\textheight]{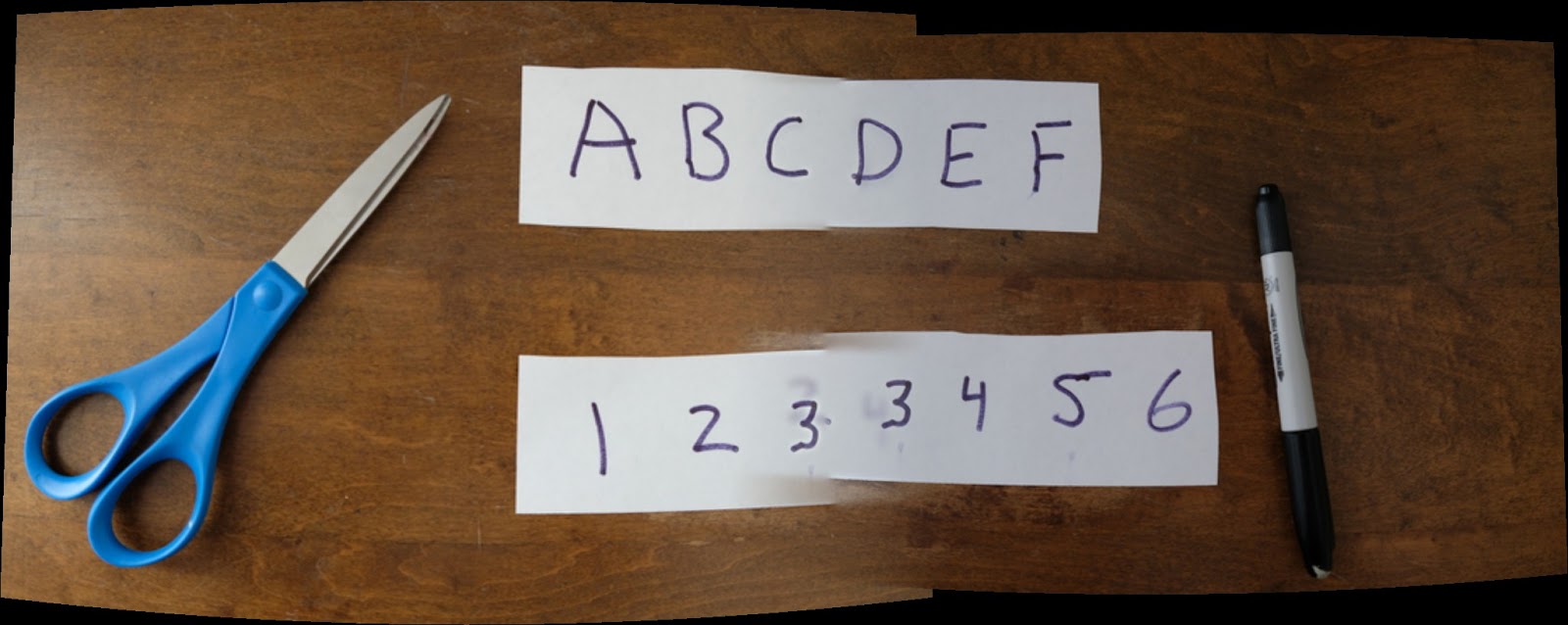}
				\caption{Autostitch~\cite{Brown_IJCV_2007}}
				\label{fig:cartoon-align-circles}
			\end{subfigure}
			\caption{
				Motion errors example.
				The strip of papers with numbers has undergone translation between input images. Our result in \ref{fig:cartoon-align-triangles} shows the use of multiple registrations. Green: the reference, Red: registration aligning the number strip, Blue: registration aligning the letter strip. Autostitch result in \ref{fig:cartoon-align-circles} has visible ghosting on the number strip.
				\label{fig:bad-stitch}
			}
		\end{centering}
	\end{figure}

	The problem of image stitching, or the creation of a panorama from a set of
	overlapping images, is a well-studied topic with widespread applications
	\cite{Kwatra:2003,Szeliski:tutorial:2006,Szeliski10}. Most modern digital cameras include a panorama creation mode, as do iPhones and Android smartphones. Google
	Street View presents the user with panoramas stitched together from
	images taken from moving vehicles, and the overhead views shown in map
	applications from Google and Microsoft are likewise stitched together from
	satellite images. Despite this ubiquity, stitching is far from
	solved. In particular, stitching algorithms often
	produce parallax errors even in a static scene with objects at different depths,
	or dynamic scene with moving objects.
	An example of motion errors is shown in Figure~\ref{fig:bad-stitch}.
	
	The stitching problem is traditionally viewed as a sequence of
	steps that are optimized independently
	\cite{Szeliski:tutorial:2006,Szeliski10}. In the first step, the algorithm computes
	a \textit{single} registration for each input image to align them to a 
	common surface.\footnote{We use the term
		registration for an arbitrary (potentially non-rigid) image
		transformation, and homography for a line-preserving image
		transformation. We will sometimes refer to the registration process as
		warping, or creating a warp.} 
	The warped images are then passed on to the
	{seam~finding} step; here the algorithm determines the registered image it should draw each pixel from.
	Finally, a {blending} procedure \cite{Perez:Poisson2003}
	is run on the composite image to correct visually unpleasant artifacts such
	as minor misalignment, or differences in color or brightness due to 
	different exposure or other camera characteristics.
	
	In this paper, we argue that currently existing methods cannot capture
	the required perspective changes for scenes with parallax or motion in
	a single registration, and that seam finding cannot compensate for
	this when the seam must pass through content-rich regions. Single
	registrations fundamentally fail to capture the background and
	foreground of a scene simultaneously. This is demonstrated in
	Figure~\ref{fig:bad-stitch}, where registering the background causes errors in the
	foreground and vice versa. Several papers \cite{Lin2016,Zhang_2014_CVPR} have addressed
	this problem by creating a single registration that is designed to produce a high
	quality stitch. However, as we will show, these still fail in cases of
	large motion or parallax due to the limitations inherent to single
	registrations. We instead propose an end-to-end approach where
	multiple candidate registrations are presented to the seam finding
	phase as alternate source images. The seam finding stage is then free to 
	choose different registrations for different regions of the composite output image.
	Note that as any registration can serve as a candidate under our
	scheme, it represents a generalization of methods that attempt to find
	a single good registration for stitching.

	Unfortunately, the classical seam finding approach \cite{Kwatra:2003} does not
	naturally work when given multiple registrations. 
	First, traditional seam finding treats each pixel from the warped image equally.
	However, by the nature of our multiple registration algorithm, each of them only
	provides a good alignment for a particular region in the image. Therefore, we need
	to consider this pixel-level alignment quality in the seam finding phase.
	Second, seam finding is performed locally by setting up an MRF that tries to place
	seams where they are not visually obvious.  Figure~\ref{fig:bad-stitch}
	illustrates a common failure; the best seam can cause objects to be
	duplicated. This issue is made worse by the use of multiple registrations.  In
	traditional image stitching, pixels come from one of two images, so in the
	worst case scenario, an object is repeated twice. However, if we use $n$
	registrations, an object can be repeated as many as $n+1$ times.  
	
	We address this issue by adding several additional terms to the MRF that
	penalize common stitching errors and encourage image accuracy.  
	Our confidence term encourages pixels
	to select their value from registrations which align nearby pixels,
	our duplication term penalizes label sets which select the same object in
	different locations from different input images, and finally our tear term penalizes
	breaking coherent regions.
	While our terms are designed to handle the challenges of multiple
	registrations, they also provide improvements to the classical
	single-registration framework.
	
	Our work can be interpreted as a layer-based approach to
	image stitching, where each registration is treated as a layer and the seam finding
	stage simultaneously solves for layer assignment and image stitching
	\cite{SergeWhatWentWhere}. Under this view, this paper represents a modest step towards explicit
	scene modeling in image stitching.
	
	
	\subsection{Motivating examples}
	
	\begin{figure}[t]
		\begin{subfigure}[b]{0.45\textwidth}
			\begin{center}
				\includegraphics[height=0.4\textwidth,angle=90]{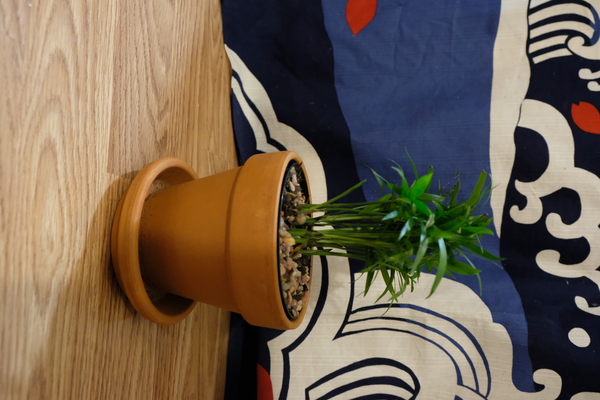}
				\includegraphics[height=0.4\textwidth,angle=90]{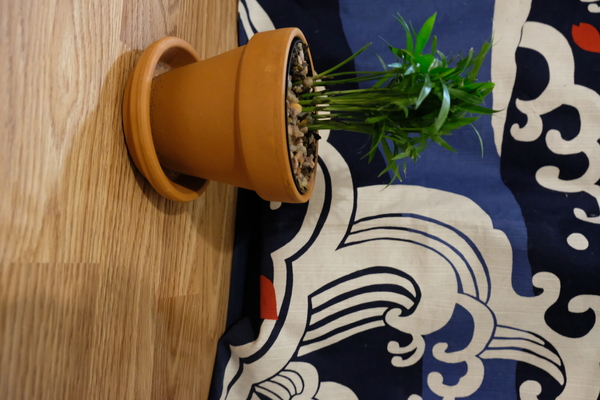}   
			\end{center}
			\caption{Input images\label{fig:plant-inputs}}
		\end{subfigure}
		\begin{subfigure}[b]{0.25\textwidth}
			\includegraphics[width=\textwidth]{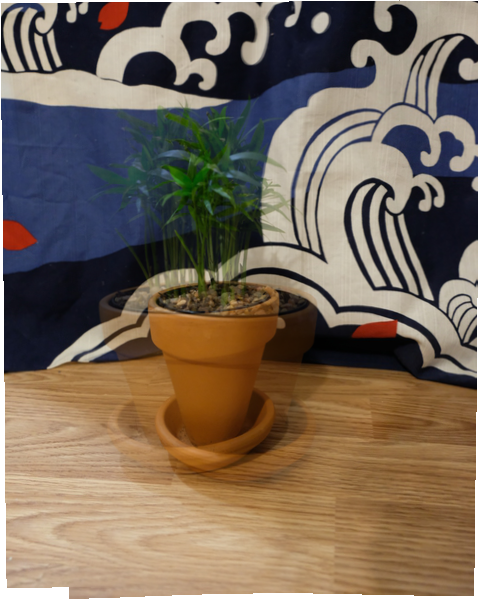}
			\caption{NIS~\cite{Chen2016} \label{fig:NIS-plant}}
		\end{subfigure}
		\begin{subfigure}[b]{0.25\textwidth}
			\sbox0{\includegraphics[width=\textwidth]{external_results/plant_1/NIS/plant2-NISwGSP3DBLEND_LINEAR.png}}
			\includegraphics[height=\ht0]{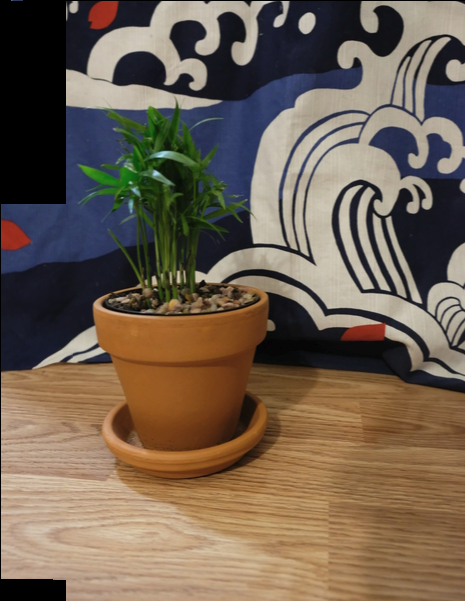}
			\caption{Our result\label{fig:plant-blend}}
		\end{subfigure}
		\caption{
			Motivating example for multiple registrations. Even the
			sophisticated single registration approach of NIS~\cite{Chen2016} gives
			severe ghosting.
			\label{fig:plant-motivation}
		}
		
		\begin{subfigure}[b]{0.35\textwidth}
			\begin{center}
				\includegraphics[width=0.45\textwidth]{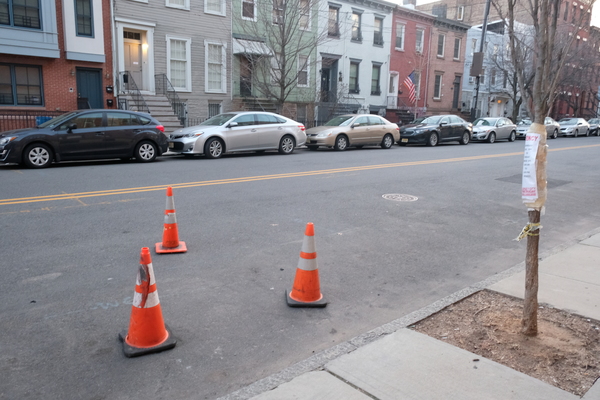}
				\includegraphics[width=0.45\textwidth]{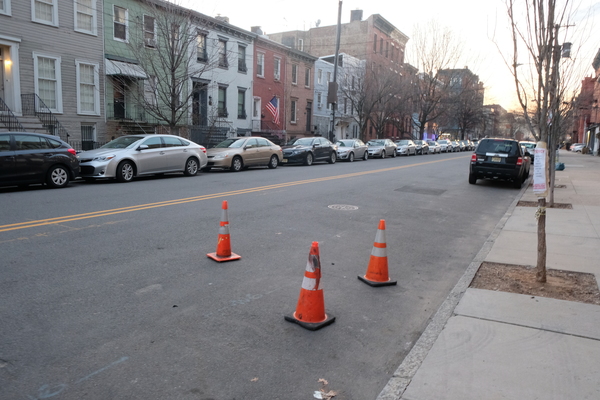}
			\end{center}
			\caption{Input images\label{fig:cones-inputs}}
		\end{subfigure}
		\begin{subfigure}[b]{0.3\textwidth}
			\includegraphics[width=\textwidth]{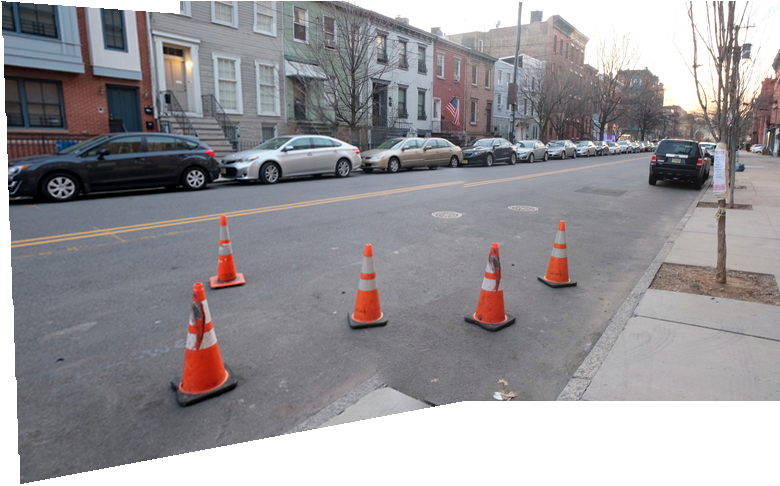}
			\caption{Photoshop~\cite{Adobe:Photomerge}\label{fig:photomerge-cones}}
		\end{subfigure}
		\begin{subfigure}[b]{0.3\textwidth}
			\includegraphics[width=\textwidth]{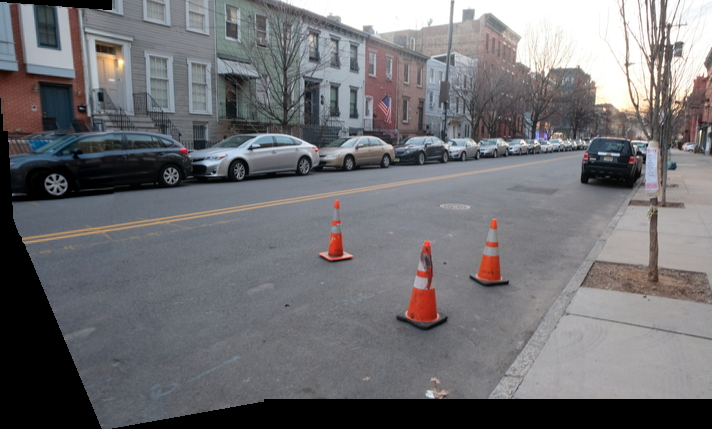}
			\caption{Our result\label{fig:cones-blend}}
		\end{subfigure}
		\caption{
			Motivating example for multiple registrations.  State of the art
			commercial packages like Adobe Photoshop~\cite{Adobe:Photomerge} duplicate the traffic cones and
			other objects.
			\label{fig:cones-motivation}
		}
	\end{figure}
	

	Figure~\ref{fig:plant-motivation} demonstrates the power of multiple registrations.
	The plant, the floor and the wall each
	undergo very distinctive motions. Our technique captures all 3 motions.
	Another challenging example is shown in Figure~\ref{fig:cones-motivation}.
	Photoshop computes a single registration to align the background buildings, which duplicates the traffic cones and the third car from left. Our technique handles all these objects at different depth correctly.

	\subsection{Problem formulation and our approach}
	We adopt the common formulation of image stitching, sometimes called
	\textit{perspective stitching} \cite{AdobeHelp:2018} or a \textit{flat
		panorama} \cite[\S 6.1]{Szeliski:tutorial:2006}, that takes one image $I_0$ as the
	reference, then warps another candidate image $I_1$ into the reference coordinate system,
	and add its content to $I_0$.
	
	Instead of proposing a single warped $\omega(I_1)$ and sending it to the seam finding phase,
	we proposed a set of warping $\omega_1(I_1), \ldots, \omega_N(I_1)$, where
	each $\omega_i(I_1)$ aligns 
	a region in $I_1$ with $I_0$. We will detail our approach for multiple registrations
	in Section~\ref{sec:multiple-regs}. Then we will formalize a
	multi-label MRF problem for seam finding. We have label set $\calL = \{0, 1,
	\ldots, N\}$, such that label $x_p = 0$ indicates
	pixel $p$ in the final stitched result will take its color value from $I_0$, and from $\omega_{x_p}(I_1)$
	when $x_p > 0$. We will get the optimal seam by minimizing the energy function $E(x)$ with
	the new proposed terms to address the challenges we introduced before.
	We will describe our seam finding energy $E(x)$ in Section~\ref{sec:seam-finding}.
	Finally, we adopt Poisson blending \cite{Perez:Poisson2003} to smooth transitions over stitching
	boundaries when generating the final result.
	
	%
	
	
	\section{Related work}
	\label{sec:related}
	
	The presence of visible seams due to parallax and other effects is a
	long-standing problem in image stitching. Traditionally there have
	been two avenues for eliminating or reducing these artifacts:
	improving registrations by allowing more degrees of freedom, or hiding
	misalignments by selecting better seams. Our algorithm can be seen as
	employing both of these strategies: the use of multiple registrations
	allows us to better tailor each registration to a particular region of
	the panorama, while our new energy terms improve the quality of the
	final seams.
	
	\subsection{Registration}
	\label{subsec:related-registration}
	
	Most previous works take a homography as a starting point and perform
	additional warping to correct any remaining
	misalignment. \cite{shum2001construction} describes a process in which
	each feature is shifted toward the average location of its matches in
	other images. The APAP algorithm divides images into grid cells and
	estimates a separate homography for each cell, with regularization
	toward a global homography \cite{zaragoza2013projective}.
	
	Instead of solving registration and seam finding independently, another line
	of work explicitly takes into account the fact that the eventual goal of the
	registration step is to produce images that can be easily stitched
	together. ANAP, for instance, can be improved by limiting perspective
	distortions in regions without overlap and otherwise regularizing to produce
	more natural-looking mosaics \cite{Lin_2015_CVPR}. Another approach
	is to confine the warping to a minimal region of the input images that is
	nevertheless large enough for seam selection and blending, which allows the
	algorithm to handle larger amounts of parallax \cite{Zhang_2014_CVPR}. Going a
	step further it is possible to interleave the registration and seam finding
	phases, as in the SEAGULL system \cite{Lin2016}. In this case, the mesh-based
	warp can be modified to optimize feature correspondences that lie close to the
	current seam.

	\subsection{Seam finding and other combination techniques}
	\label{subsec:related-seam-finding}
	
	The seam finding phase requires determining, for each pixel, which of
	the two source images contributes its color. \cite{Kwatra:2003}
	observed that this problem can be naturally formulated as a Markov
	Random Field and solved via graph cuts. This approach tends to give
	strong results, and the graph cuts method in particular often produces
	energies that are within a few percent of the global minimum
	\cite{SZSVKATR:PAMI08}. Further work in this area has focused on
	taking into account the presence of edges or color gradients in the
	energy function in order to avoid visible discontinuities.
	\cite{Agarwala:2004:IDP:1015706.1015718}.
	
	An alternative to seam finding is the use of a a multi-band blending
	\cite{Burt_TOG_83} phase immediately after registration \cite{Brown_IJCV_2007}. This
	step blends low frequencies over a large spatial range and high
	frequencies over a short range to minimize artifacts.
	
	
	

	\subsection{Comparison to our technique}
	\label{subsec:related-comparisons}
	
	Our work clearly generalizes the line of work that optimizes a single
	registration, as this arises as a special case when only one candidate
	warp is used. More usefully, existing registration methods can serve
	as candidate generators in our technique. A single registration
	algorithm can propose multiple candidates when run with different
	parameters, or in the case of a randomized algorithm, such as RANSAC,
	run several times.
	
	Similarly, our algorithm can be viewed as implicitly defining a single
	registration, given at each pixel by the warp $\warp_i$ associated with the
	candidate registration from which the pixel was drawn in the final
	output.
	In theory, this piecewise defined warp is sufficient to obtain the results
	reported here, but in practice, finding it is difficult.
	Previous work along these lines has focused on iterative schemes in order to compute
	the varying warps that are required in different regions of the image
	\cite{Chen2016,Lin_2015_CVPR}, but this is in general a very computationally
	challenging problem and the warping techniques used may not be sufficient to produce
	a good final results. Our technique allows multiple simple registrations to be
	used instead.

	\section{Our multiple registration approach}
	
	We use a classic three stage image stitching pipeline, composed of
	registration, seam finding, and blending phases \cite{Szeliski:tutorial:2006,Szeliski10}.
	
	In the registration phase, we propose multiple registrations, each of
	which attempts to register some part of one of the images with the
	other. In contrast to previous methods, which only pass a single
	proposed registration to the seam finding stage, our approach allows
	all of these proposed registrations to be used. Note that in this phase it is
	important that the set of registrations we propose be diverse.
	
	In the seam finding stage, we solve an MRF inference problem to find
	the best way to stitch together the various proposals.  
	We observed that using traditional MRF energy to stitch multiple registrations naively
	generated poor results,
	due to the reasons we mentioned in Section~\ref{sec:introduction}.
	To address these challenges, we propose the improved MRF energy 
	by adding (1) a new data term that describes our
	confidence between different warping proposals at pixel $p$ and 
	(2) several new smoothness terms which attempt to prevent duplication or tearing. 
	Although this new energy is proposed primarily for the stitching problem
	with multiple registrations, it addresses problems observed in the traditional approach (single registration) as well 
	and provides marked improvements in final panorama quality in either framework.
	
	Finally, we adopt Poisson blending \cite{Perez:Poisson2003} to smooth transitions over stitching
	boundaries when generating the final result.
	
	\subsection{Generating multiple registrations}
	\label{sec:multiple-regs}
	
	There are two common categories of registration methods \cite{Szeliski10}:
	\emph{global} transformations, implied by a single motion model over the whole
	image, such as a homography; and \emph{spatially-varying} transformations,
	implicitly described by a non-uniform mesh. The candidate registrations we
	produce are spatially-varying non-rigid transformations.  Similar to
	\cite{Zhang_2014_CVPR}, we first obtain a homography that matches some
	part of the image well and then refine its mesh representation.
	
	We have a 3 step process: homography finding, filtering, and refinement.  In
	the homography finding step, we generate candidate homographies by running
	RANSAC on the set of sparse correspondences between features obtained from the
	two input images. We ensure that the set of homographies is diverse by a
	filtering step, which removes poor quality homographies and duplicates.  In
	the refinement step, we solve a quadratic program (QP) to obtain an improved
	local warping mesh for each of the homographies that pass the filtering step.
	
	\noindent\textbf{Homography finding step.} Given two input images $I_0$ and $I_1$, we
	first compute a set of sparse correspondences $C=\{ (p^0_1, p^1_1), \ldots, (p^0_n,
	p^1_n) \}$, where each $p^0_i \in I_0$, $p^1_i \in I_1$ and $(p^0_i,p^1_i)$ is a pair of
	matched pixels. We run $\tau_H$ iterations of a modified RANSAC algorithm to generate
	a set of potential homographies $\mathcal{H}$.  In each iteration $t$, we
	randomly choose a pixel $p$ and consider correspondences within a distance
	$r_H$; if there are enough nearby correspondences to allow us to estimate a
	homography $H_t$ we add this to our set of candidates.  The homography $H_t$
	is estimated using least median of squares as implemented in
	OpenCV~\cite{opencv_library}.
	
	\noindent\textbf{Filtering step.} In order to simplify the seam finding step, it is
	desirable to limit the number of candidate homographies. We employ two strategies to
	achieve this: \emph{screening}, which removes homographies from
	consideration as soon as they are found, and \emph{deduplication},
	which runs on the full set of homographies that remain after
	screening.
	
	The screening procedure eliminates two kinds of
	homographies: those that are unlikely to give rise to realistic
	images, and those that are too close to the identity transformation to
	be useful in the final result. Homographies of the first type are
	eliminated by considering two properties: (1) whether the difference
	between a \emph{similarity} motion that is obtained from the same set
	of seed points exceeds a fixed threshold \cite[\S
	3.2.1]{Zhang_2014_CVPR}, and (2) whether the magnitude of the
	scaling parameters of the homography exceed a (different) fixed
	threshold. The intuition is that real world perspective
	changes are often close to similarities, and stitchable
	images are likely to be close to each other in scale. Homographies
	that are too close to $I$ are eliminated by checking whether the
	overlap between the area covered by the original image and the area
	covered by the transformed image exceeds 95\%. Finally, we reject homographies
	where either diagonal is shorter than half the length of the diagonal of the
	original image.
	
	
	To determine the set of homographies that are near-duplicates of each
	other and of which all but one can therefore be safely discarded, we
	compute a set of inlier correspondences $D_t$ for each $H_t$ that
	passes screening. $D_t$ is constructed iteratively, starting with all
	correspondences $(p^0_i, p^1_i) \in C_t'$, where $C_t'$ is the subset
	of seed points that were chosen in iteration $t$ for which the
	reprojection error is below a threshold $T_H$. Correspondences
	containing points that lie within a distance $r_D$ of some point
	already in $D_t$ are then added until a fixpoint is reached. This step
	is a generalization of the strategy introduced in \cite[\S
	3.2.1]{Zhang_2014_CVPR}.
	
	Given the sets $D_t$ computed for each $H_t$, we define a
	\emph{similarity measure} between homographies $\text{sim}(H_a, H_b) =
	\cos(V_a, V_b)$, where $\cos$ represents the cosine distance and $V_a$
	the $0$-$1$ indicator vector for $D_t$. Homographies are then considered
	in descending order of $|D_t|$ and added to the set $\mathcal{H}$ if
	their similarity to all the elements that have already been added to
	the set is below a threshold $\theta_H$. We also enforce an upper
	limit $N_H$ on the number of homographies considered, terminating the
	procedure early when this limit is reached.
	
	\noindent\textbf{Refinement step.}
	Our final step is motivated by the observation that our process sometimes
	produces homographies that cause reprojection errors of several pixels. This
	may occur even for large planar objects, such as the side of a building, which
	should be fit exactly by a homography.
	We make a final adjustment to our homography, then add spatial
	variation.
	
	To adjust the homography, we define an objective function 
	\begin{math}
	f(H) = \sum_{c_i \in C} S(e_{c_i;H}),
	\end{math}
	where $e_{c_i;H}$ is the reprojection error of correspondence $c_i$
	under $H$, and $S$ is a smoothing function
	\begin{math}
	S(t) = 1 - \frac{1}{1+\exp(-(T_H-t))}.
	\end{math}
	To generate a \emph{refined homography} $\hat H_i$ from an input $H_i$, we
	minimize $f$ using Ceres~\cite{ceres-solver}, initializing with $H_i$.  The resulting
	$\hat H_i$ is a better-fitting homography that is in some sense near $H_i$.  The
	smoothing function $S$ is designed to provide gradient in the right direction
	for correspondences that are close to being inliers while ignoring those that
	are outliers either because they are incorrect matches or because they are
	better explained by some other homography.
	
	The homographies $\hat H_i \in \mathcal{H}$ often do an imperfect job of
	aligning $I_0$ and $I_1$ in regions that are only mostly flat. In order to
	address this, we compute a finer-grained non-rigid registration $\warp_i$ for
	each $\hat H_i$ using a content-preserving warp (CPW) technique that is better
	able to capture the transformation between the two
	images~\cite{Liu:SIGGRAPH09}.  We start from a uniform grid mesh $M_i$ drawn
	over $\hat H_i(I_1)$, and attempt to use CPW to get a new mesh $\hat M_i$ to capture
	fine-grained local variations between $I_0$ and $H_i(I_1)$. 
	
	Finally, we denote by $\omega_i(I_1)$ the warped candidate image $I_1$ with
	$\hat M_i$ applied.

	\subsection{Improved MRF energy for seam finding}
	\label{sec:seam-finding}
	
	The final output of the registration stage is a set of proposed warps
	$\{\warp_i(I_1) \}, (i = 1, 2, \ldots, N)$. For notational simplicity, we write $\{ I^S_i
	\}$ where  $I^S_0 = I_0$, $I^S_i = \omega_i(I_1)$ are the source images in
	the seam finding stage. These images are used to
	set up a Markov Random Field (MRF) inference problem, to decide
	how to combine regions of the different
	images in order to obtain the final stitched image.  The label set for
	this MRF problem is given by $\calL = \{0, 1, \ldots, N\}$, and
	its purpose is to assign a label $x_p \in \calL$ to each pixel $p$ in the
	stitching space, which indicates that the value of that pixel is copied from $I^S_{x_p}$. 
	
	\def\oldE{E^{\text{old}}}
	
	It would be natural to expect that we can use the standard MRF stitching energy
	function $\oldE(x) = \sum_{p} \oldE_m(x_p) + \sum_{p, q \in \calN} \oldE_s(x_p, x_q)$ 
	introduced by \cite{Kwatra:2003} (where $\calN$ is the 4-adjacent neighbors).
	However, we observed that this energy function is not suitible for the case of multiple registrations.
	
	In this formulation, the data term $\oldE_m(x_p) = 0$ when pixel $p$ has a
	valid color value in $I^S_{x_p}$, and $\lambda_m$ otherwise. This means we will
	impose a penalty $\lambda_m$ for out-of-mask pixels but treat all the
	inside-mask pixels equally (they all have cost 0). However, we found that even
	state-of-the-art single-registration algorithms~\cite{Lin2016,Zhang_2014_CVPR},
	cannot align every single pixel.  In contrast, our multiple registrations are
	designed to only capture a single region with each warp.  We propose a new
	mask data term for multiple registrations and a warp data term to address this
	problem.
	
	The traditional smoothness term is $\oldE_s(x_p, x_q) = \lambda_s (\|
	I^S_{x_p}(p) - I^S_{x_q}(p)\| + \| I^S_{x_p}(q) - I^S_{x_q}(q)\|)$ when $x_p
	\ne x_q$, and 0 otherwise.  It only enforces local similarity across the
	stitching seam to make it less visible, without any other global constraints.
	Note that there are a number of nice extensions to this basic idea that improve
	the smoothness term; for example~\cite[p. 62]{Szeliski:tutorial:2006} describes
	several ways to pick better seams and avoid tearing. However, we may still
	duplicate content in the stitching result with a single registration due to
	parallax or motion. This problem can be more serious with multiple
	registrations since we may duplicate content $N+1$ times instead of just twice.
	Therefore, we propose a new pairwise term to explicitly penalize duplications.
	
	In sum, we compute the optimal seam by minimizing the energy function $E(x) =
	\sum_p E_m(x_p) + \sum_p E_w(x_p) + \sum_{p, q \in \calN} E_s(x_p, x_q) +
	E_d(x)$ using expansion moves~\cite{BVZ:PAMI01}. We now describe our mask data
	term $E_m$, warp data term $E_w$, smoothness term $E_s$ and duplication term
	$E_d$ in turn.

	\noindent \textbf{Mask data term for multiple registrations.}
	There is an immediate issue with the standard mask-based data term in the
	presence of multiple registrations. When one input is significantly
	larger than the others, the MRF will choose this warping for pixels
	where its mask is 1 and the other warping masks are 0. Worse,
	since the MRF itself imposes spatial coherence, this choice of input will
	be propagated to other parts of the image.
	
	We handle this situation conservatively, by imposing a mask penalty $\lambda_m$ on pixels
	that are not in the intersection of all the candidate warpings $\bigcap_i \warp_i(I_1)$
	when assigning them to a candidate image (i.e., $x_p \ne 0$).
	Pixels that lie inside the reference image ($x_p = 0$) are handled normally,
	in that they have no mask penalty with the reference image mask and $\lambda_m$
	mask penalty out of the mask.
	Note that this mask penalty is a soft constraint: pixels outside of the intersection $\bigcap_i \warp_i(I_1)$
	can be assigned an intensity from a candidate image, if it is promising enough by our other criteria.
	
	Formally we can write our mask data term as
	\begin{equation}
	E_m(x_p) = \begin{cases}
	\lambda_m \left(1 - \mathsf{mask}_0(p) \right), & x_p = 0, \\
	\lambda_m \left(1-\prod_{i=1}^N\mathsf{mask}_i(p) \right), & x_p \ne 0,
	\end{cases}
	\end{equation}
	where $\mathsf{mask}_i(p) = 1$ indicates $I^S_i$ has a valid pixel at $p$, $\mathsf{mask}_i(p) = 0$ otherwise.
	
	\noindent \textbf{Warp data term.}
	In the presence of multiple registrations, we need a data term that makes
	significant distinctions among different proposed warps. There are two natural
	ways to determine whether a particular warp $\warp$ is a good choice at the pixel
	$p$. First, we can determine how confident we are that $\warp$ actually
	represents the motion of the scene at $p$. Second, for pixels in the reference
	image, we can check intensity/color similarity between $I_0(p)$ and
	$\warp(I_1)(p)$. 
	
	
	Since our warp is computed using features and RANSAC, we can identify inlier
	feature points in $\omega_i(I_1)$ when the reprojection error is smaller than a
	parameter $T_H$.  Denoting these inliers as $\calI_i$, we place a Gaussian
	weight $G(.)$ on each inlier, and define motion quality for pixel $p$ in
	$I^S_i$ as $Q^i_m(p) = \sum_{q \in \calI_i} G(\| p - q\|)$.  This makes pixels
	closer to inliers have greater confidence in the warp.
	
	For color similarity we use the $L_2$ distance between the local patch around
	pixel $p$ in the reference $I^S_0$ and our warped image $I^S_i$:
	$Q^i_c(p) = \sum_{q \in \calB_r(p)} \| I^S_0(p) - I^S_i(p) \|$, where $\calB_r(p)$
	is the set of pixels within distance $r$ to pixel $p$. So pixels with better image
	content alignment become more confident in the warp.
	
	Putting them together, we have $e^i_w(p) = -Q^i_m(p) + \lambda_c Q^i_c(p)$
	to be our quality score for pixel $p$ for warp $\omega_i$
	(lower means better, since we want to minimize the energy). Then we have a
	normalized score $\hat e^i_w(p) \in [-1, 1]$ per warped image, and define
	the warp data term as: $E_w(x_p) = \lambda_w \hat e^{x_p}_w(p)$ when
	$x_p \ne 0$, and $E_w(x_p) = 0$ otherwise.
	

	\noindent \textbf{Smoothness terms.}
	We adopt some standard smoothness terms used in state-of-the-art MRF stitching.
	Following \cite{Szeliski:tutorial:2006,Szeliski10} these terms include:
	\begin{enumerate}
		\item the color-based seam penalty (introduced in
		\cite{Kwatra:2003,Agarwala:2004:IDP:1015706.1015718})
		for local patches to encourage seams that introduce invisible
		transitions between source images,
		\item the edge-based seam penalty introduced in \cite{Agarwala:2004:IDP:1015706.1015718}
		to discourage the seam from cutting through edges, hence reduce the ``tearing''
		artifacts where only some part of an object appears in the stitched result,
		\item a Potts term to encourage local label consistency.
	\end{enumerate}
	
	
	\begin{figure}[t]
		\setlength{\belowcaptionskip}{0pt}
		\begin{centering}
			\begin{subfigure}[t]{0.24\textwidth}
				\includegraphics[width=\textwidth]{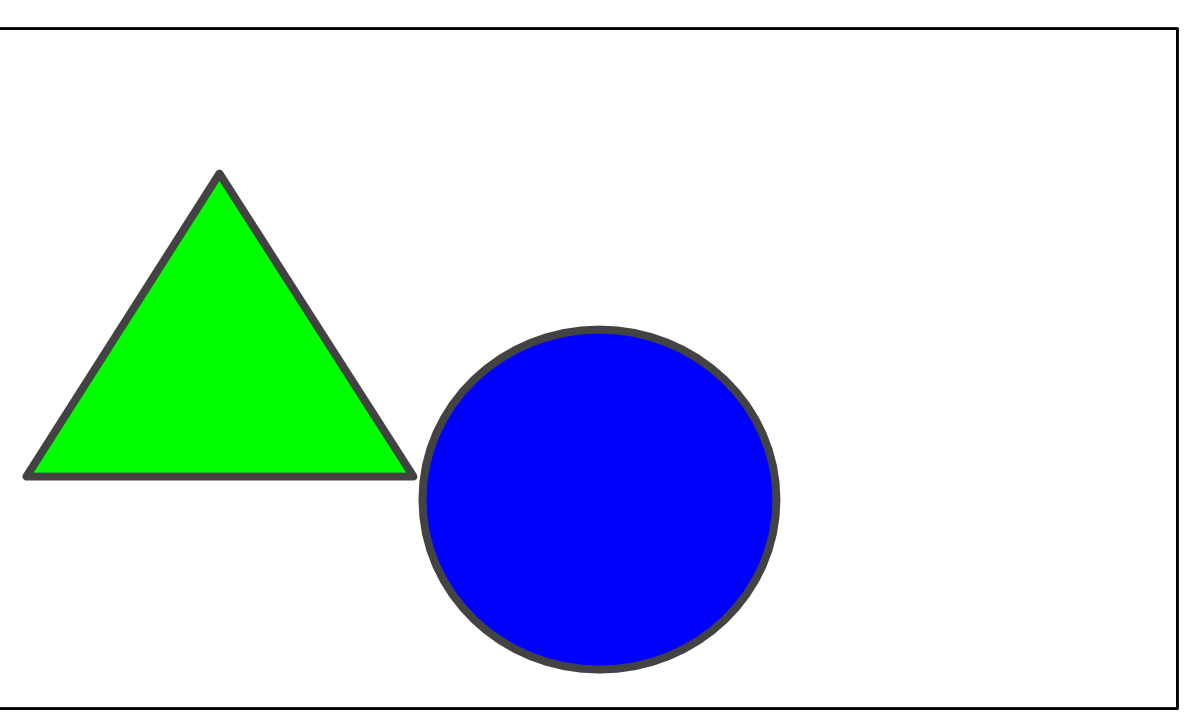}
				\caption{Candidate image}
			\end{subfigure}
			\begin{subfigure}[t]{0.24\textwidth}
				\includegraphics[width=\textwidth]{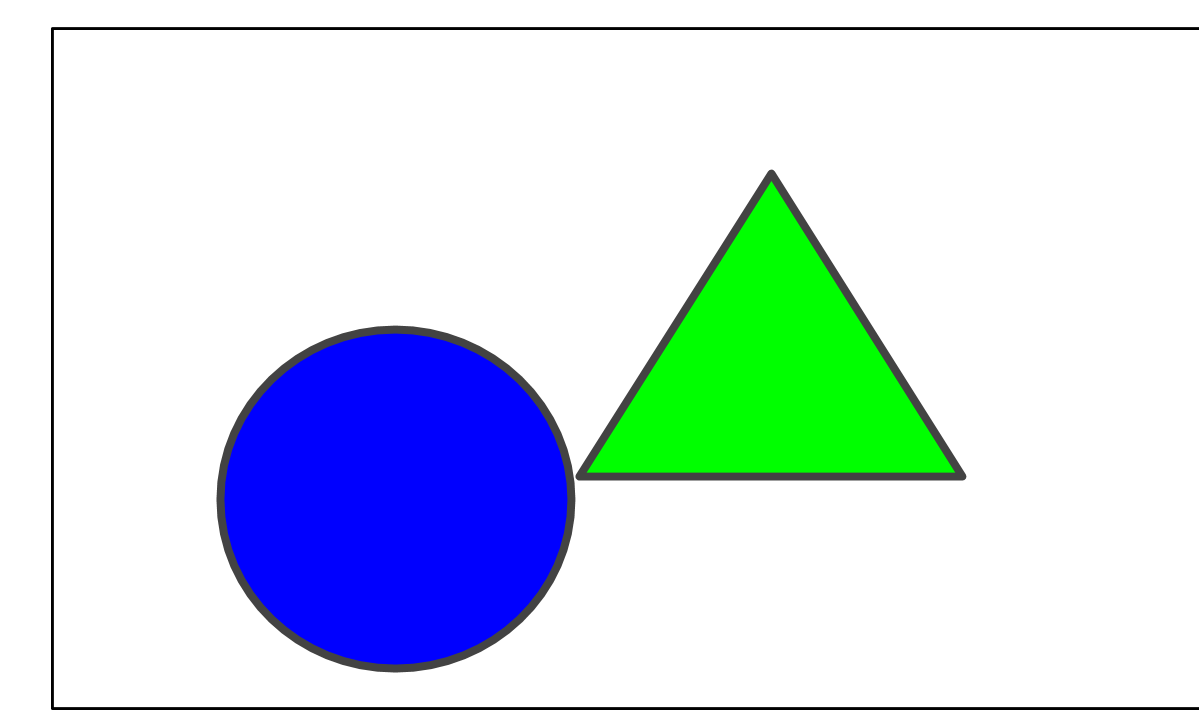}
				\caption{Reference image}
			\end{subfigure}
			\begin{subfigure}[t]{0.24\textwidth}
				\includegraphics[width=\textwidth]{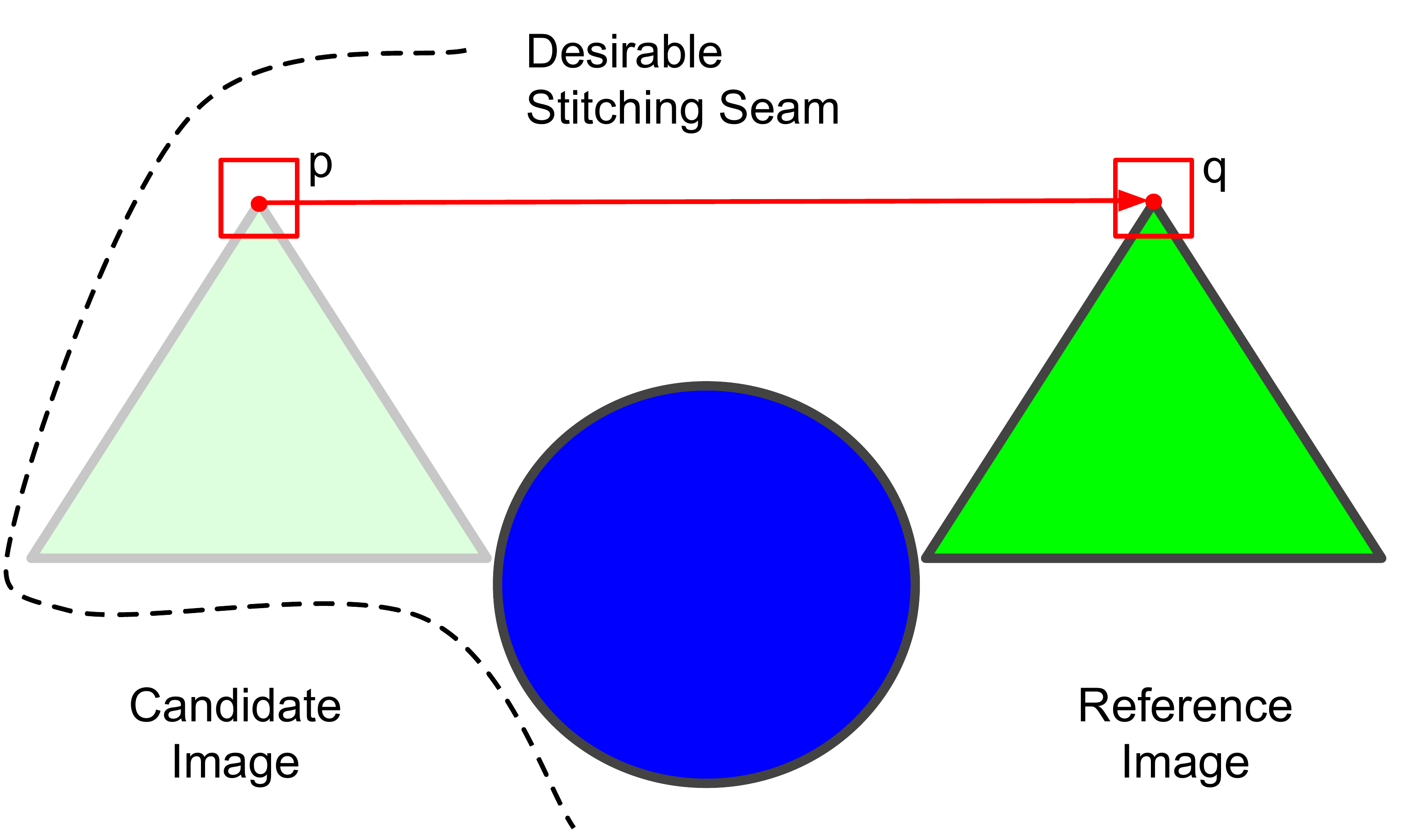}
				\caption{Good stitch\label{fig:bad_stitching}}
			\end{subfigure}
			\begin{subfigure}[t]{0.24\textwidth}
				\includegraphics[width=\textwidth]{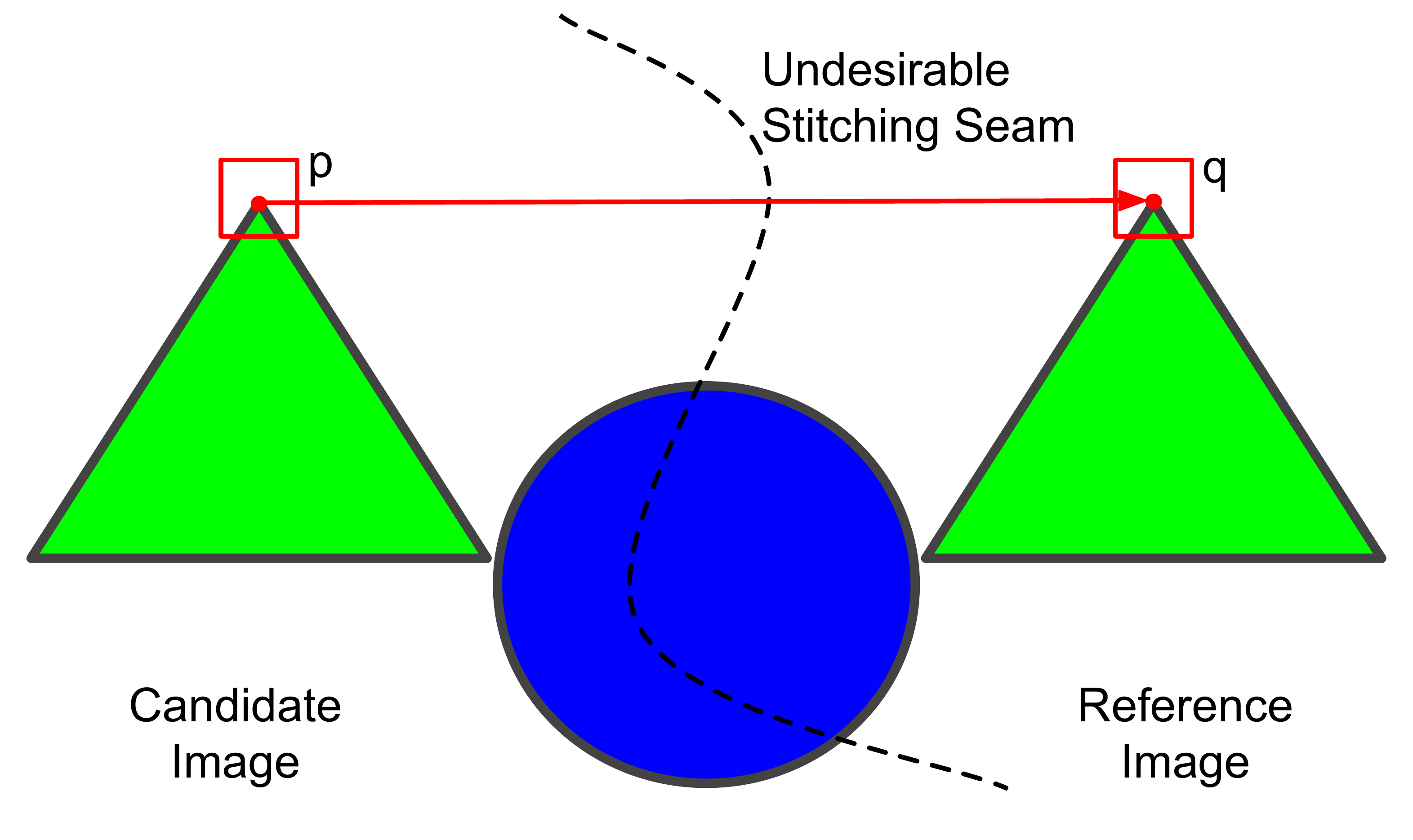}
				\caption{Bad stitch\label{fig:good_stitching}}
			\end{subfigure}
			\caption{
				Illustration of the duplication term. Figure~\ref{fig:bad_stitching} provides a bad
				stitching result with the green triangles duplicated.
				The feature point correspondence between pixel $p$ and
				$q$ suggests duplication, and we introduce a term
				which penalizes this scenario.
				\label{fig:duplication_term}
			}
		\end{centering}
	\end{figure}
	
	\noindent \textbf{Duplication avoidance term.} For stitching tasks with large
	parallax or motion, it is easy to duplicate scene content in the stitching result.  We
	address this issue by explicitly formalizing a duplication avoidance term in our energy.
	If pixel $p$ from the reference image $I^S_0$ and $q$ from the candidate image
	$I^S_i$ form a true correspondence, then they refer to the same point (i.e., scene
	element) in the real world. Therefore, we penalize a labeling that contains both
	of them (i.e., $x_p = 0, x_q = i$), as shown in Figure~\ref{fig:duplication_term}. Since
	our correspondence is sparse, we also apply this idea to the local
	region within a radius $r$ of pixels $p$ and $q$.  We reweight the
	penalty by a Gaussian $G$ since the farther away we are from these
	corresponding pixels, the more uncertain the correspondence.
	
	Formally, our duplication term $E_c$ is defined as
	
	\begin{equation}
	E_d(x) = \lambda_d \sum_{i = 1}^N \sum_{(p, q) \in \calC_i} \sum_{\delta
		\in \calB_r} e_r(x_{p + \delta}, x_{q + \delta}; \delta, i) 
	\end{equation}
	where $\calC_i$ is the pixel correspondence between $I^S_0$ and $I^S_i$, 
	and $\calB_r = \{(dx, dy) \in \calI^2 \mid \| (dx, dy)\| \le r\}$ is a box of
	radius $r$. $e_r(x_{p+\delta}, x_{q+\delta}; \delta, i) = G(\| \delta \|)$ when 
	$x_p = 0, x_q = i$, and 0 otherwise.

	\newcommand{\resultsfigure}[4]{
		\begin{center}
			\begin{minipage}{0.31\textwidth}
				\setlength{\abovecaptionskip}{1pt}
				\setlength{\belowcaptionskip}{1pt}
				\includegraphics[width=0.48\textwidth]{input_images/#1/reference.jpg}
				\includegraphics[width=0.48\textwidth]{input_images/#1/candidate.jpg}
				\captionof*{figure}{(a) Inputs}
			\end{minipage}
			\begin{minipage}{0.31\textwidth}
				\includegraphics[width=\textwidth]{external_results/#1/#3}
				\captionof*{figure}{(b) #4}
			\end{minipage}
			\begin{minipage}{0.31\textwidth}
				\includegraphics[width=\textwidth]{data_results_2018_03_05/#1/ours_blend_expand_crop.png}
				\captionof*{figure}{(c) Our result}
			\end{minipage}
			\captionof{figure}{\label{fig:#1}#2}
		\end{center}
	}
	
	\newcommand{\resultsfigurechen}[4]{
		\begin{center}
			\begin{subfigure}{0.31\textwidth}
				\includegraphics[width=0.48\textwidth]{input_images/#1/reference.jpg}
				\includegraphics[width=0.48\textwidth]{input_images/#1/candidate.jpg}
				\caption{Input images}
			\end{subfigure}
			\begin{subfigure}{0.31\textwidth}
				\includegraphics[width=\textwidth]{external_results/#1/#3}
				\caption{#4}
			\end{subfigure}
			\begin{subfigure}{0.31\textwidth}
				\includegraphics[width=\textwidth]{data_results_2018_03_05/#1/ours_blend_expand_crop.png}
				\caption{Our result}
			\end{subfigure}
			\caption{\label{fig:#1}#2}
		\end{center}
	}
	
	\newcommand{\bigresultsfigure}[2]{
		\begin{center}
			\begin{subfigure}{0.31\textwidth}
				\includegraphics[width=0.48\textwidth]{input_images/#1/reference.jpg}
				\includegraphics[width=0.48\textwidth]{input_images/#1/candidate.jpg}
				\caption{Input images}
			\end{subfigure}
			\begin{subfigure}{0.31\textwidth}
				\includegraphics[width=\textwidth]{external_results/#1/apap_blend.png}
				\caption{\label{fig:#1_apap} APAP~\cite{zaragoza2013projective}}
			\end{subfigure}
			\begin{subfigure}{0.31\textwidth}
				\includegraphics[width=\textwidth]{external_results/#1/photomerge.png}
				\caption{\label{fig:#1_photomerge} Photoshop~\cite{Adobe:Photomerge}}
			\end{subfigure}
			\begin{subfigure}{0.31\textwidth}
				\includegraphics[width=\textwidth]{external_results/#1/nis.png}
				\caption{\label{fig:#1_nis} NIS~\cite{Chen2016}}
			\end{subfigure}
			\begin{subfigure}{0.31\textwidth}
				\includegraphics[width=\textwidth]{data_results_2018_03_05/#1/ours_blend_expand_crop.png}
				\caption{\label{fig:#1_ours} Our result}
			\end{subfigure}
			\caption{\label{fig:#1}#2}
		\end{center}
	}
	
	\section{Experimental results and implementation details}
	
	\noindent \textbf{Experimental setup.} Our goal is to perform stitching on images whose degree of parallax and motion
	causes previous methods to fail.  Ideally, there would be a standard dataset
	of images that are too difficult to stitch, along with an evaluation metric.
	Unfortunately this is not the case, in part due to the difficulty of defining ground
	truth for image stitching.  We therefore had to rely on collecting challenging
	imagery ourselves, though we found one appropriate example (Figure~\ref{fig:ski_left})
	whose stitching failures were widely shared on social media.
	
	We implemented or obtained code for a number of alternative methods, as
	detailed below, and ran them on all of our examples, along with our technique
	using a single parameter setting. Since our images are so challenging, it was
	not uncommon for a competing method to return no output (``failing to
	stitch'').  In the entire corpus of images we examined, we found numerous
	cases where competing techniques produced dramatic artifacts, while our
	algorithm had minimal if any errors.  We have not found any example images
	where our technique produces dramatic artifacts and a competitor does
	not. However, we found a few less challenging images that are well handled by
	competitors but where we produce small artifacts.  These examples, along
	with other data, images, and additional material omitted here are
	available online,\footnote{See \url{https://sites.google.com/view/oois-eccv18}.} for reasons of space we
	focus here on images that provide useful insight. However, the images
	included here are representative of the performance we have observed on the
	entire corpus of challenging images we collected.
	
	We follow the experimental setup
	of~\cite{Zhang_2014_CVPR}, who (very much like our work) describe a stitching
	approach that can handle images with too much parallax for previous
	techniques.   The strongest overall competitor turns out to be
	Adobe Photoshop 2018's stitcher Photomerge~\cite{Adobe:Photomerge}.
	While experimental results reported in \cite{Zhang_2014_CVPR} compare
	their algorithm with Photoshop 2014, the 2018 version is substantially better,
	and does an excellent job of stitching images with too many motions for any
	other competing methods. Therefore, we take Photoshop's failing on a dataset
	as a signal that that dataset is particularly challenging; in this section, we
	show several examples in this section where we successfully stitch such
	datasets. In addition to Photoshop we downloaded and ran APAP
	\cite{zaragoza2013projective}, Autostitch \cite{Brown_IJCV_2007}, and NIS
	\cite{Chen2016}. To produce stitching results from APAP we follow the
	approach of \cite{Zhang_2014_CVPR}, who extended APAP with seam-finding.
	Results from all methods are shown in Figure~\ref{fig:graffiti_car_1}
	and~\ref{fig:cars_1}.
	
	
	\noindent \textbf{Implementation details.} For feature extraction and matching, we used
	DeepMatch~\cite{Weinzaepfel_2013_ICCV}.  The associated DeepFlow solver was
	used to generate flows for the optical flow-based warping.  We used the Ceres
	solver \cite{ceres-solver} for the QP problems that arose when generating
	multiple registrations, as discussed in section~\ref{sec:multiple-regs}.
	
	\noindent \textbf{Visual evaluation.} Following \cite{Zhang_2014_CVPR} we review
	several images from our test set and highlight the strengths and weaknesses of
	our technique, as well as those of various methods from the literature. All of
	our results are shown for a single set of parameters.
	
	We observed two classes of stitching errors: {warping errors}, where the
	algorithm fails to generate any candidate image that is well-aligned with the
	reference image; and {stitching errors}, where the MRF does not produce good
	output despite the presence of good candidate warps. An example of our
	technique making a 
	\emph{warping error} is shown in Figure~\ref{fig:graffiti_car_1_ours}, where no warp found
	by our algorithm continues the parking stall line, causing a visible seam. An
	example of a \emph{stitching error} is given in
	Figure~\ref{fig:cars_1_ours}, where the remainder of the car's wheel
	is available in the warp from which our mosaic draws the front wheelwell.
	Errors may manifest as a number of different kinds of artifacts, such as:
	tearing (e.g., the arm in Figure~\ref{fig:ski_left}b); wrong
	perspective (e.g., the tan background building in
	Figure~\ref{fig:graffiti_car_1_apap}); or duplication (e.g., the stop sign
	in~\ref{fig:stop_sign_1}b), ghosting (e.g., the bollards in
	Figure~\ref{fig:biker_mural_1}b), or omission (e.g., the front door
	of the car in Figure~\ref{fig:cars_1_photomerge}) of scene content.
	
	\begin{figure}[!htp]
		\setlength{\abovecaptionskip}{0pt}
		\setlength{\belowcaptionskip}{0pt}
		\resultsfigurechen{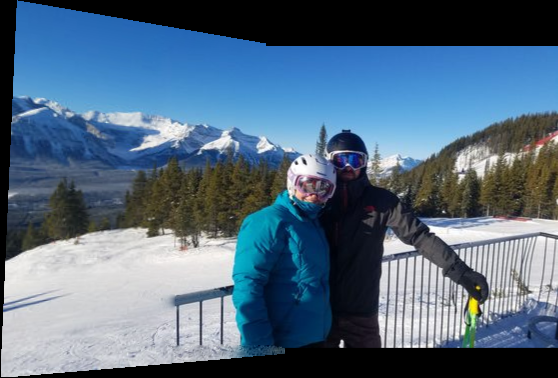}{``Ski'' dataset. Photoshop tears the people and the fence. Our stitch
			has the fence stop abruptly but keeps the
			people in place. Note that the candidate provides no information that allows us to extend the fence.}{photomerge.png}{Photoshop~\cite{Adobe:Photomerge}}
		
		\resultsfigurechen{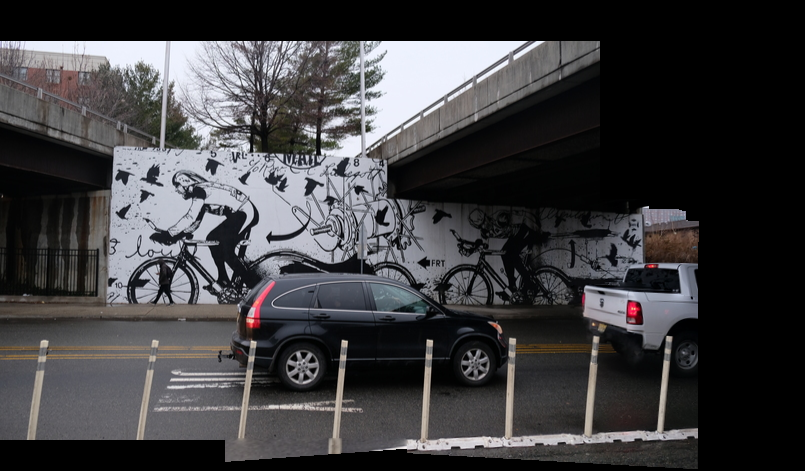}{``Bike Mural'' dataset. Autostitch has ghosting on the car, bridge, and poles. Our
			algorithm shortens the truck and deletes a pole, but has no perceptible
			ghosting or tearing of the objects.}{autostitch.jpg}{Autostitch~\cite{Brown_IJCV_2007}}
		
		\resultsfigurechen{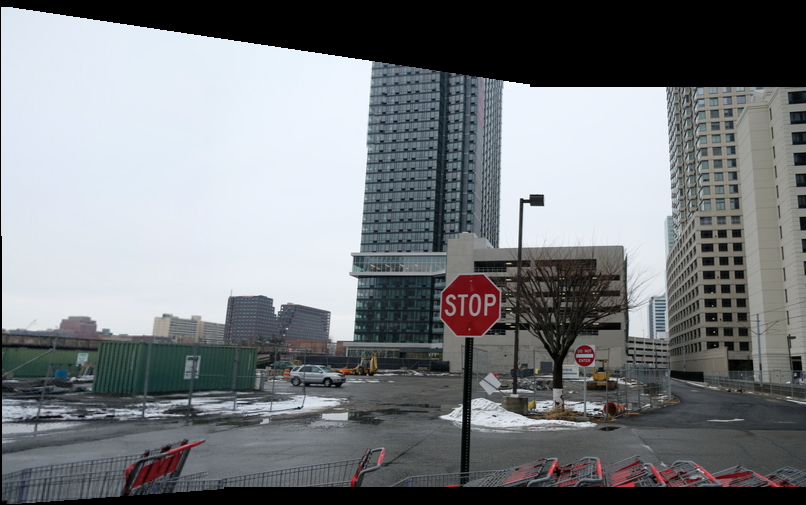}{``Stop Sign'' dataset. Photoshop duplicates the stop sign. Of all the implementations we tried, ours is the
			only visually plausible result, successfully avoiding duplicating the
			foreground.  }{photomerge.png}{Photoshop~\cite{Adobe:Photomerge}}
		
		\resultsfigurechen{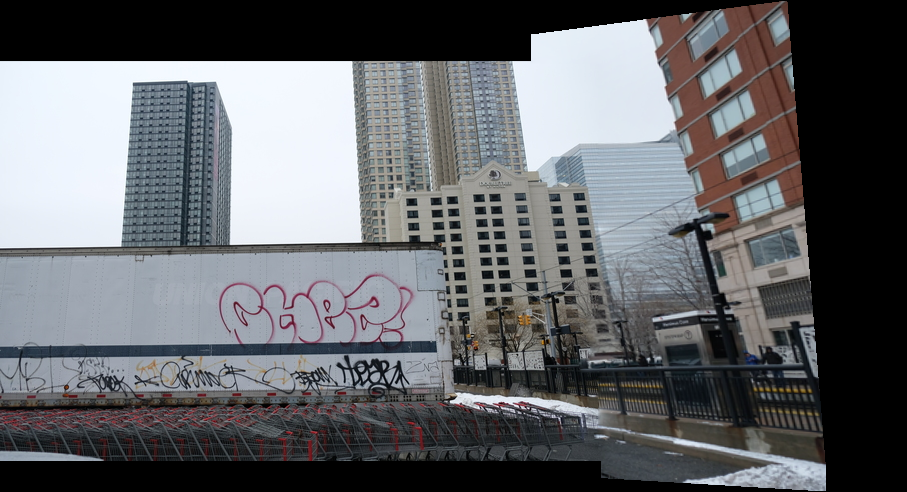}{``Graffiti-Building''
			dataset. APAP deletes significant
			amounts of red graffiti, and introduces noticable curvature. Our result does not produce tearing, ghosting, or
			duplication.}{apap_blend.png}{APAP~\cite{zaragoza2013projective}}
	\end{figure}%
	
	\begin{figure}[!htp]
		\setlength{\abovecaptionskip}{0pt}
		\setlength{\belowcaptionskip}{0pt}
		\bigresultsfigure{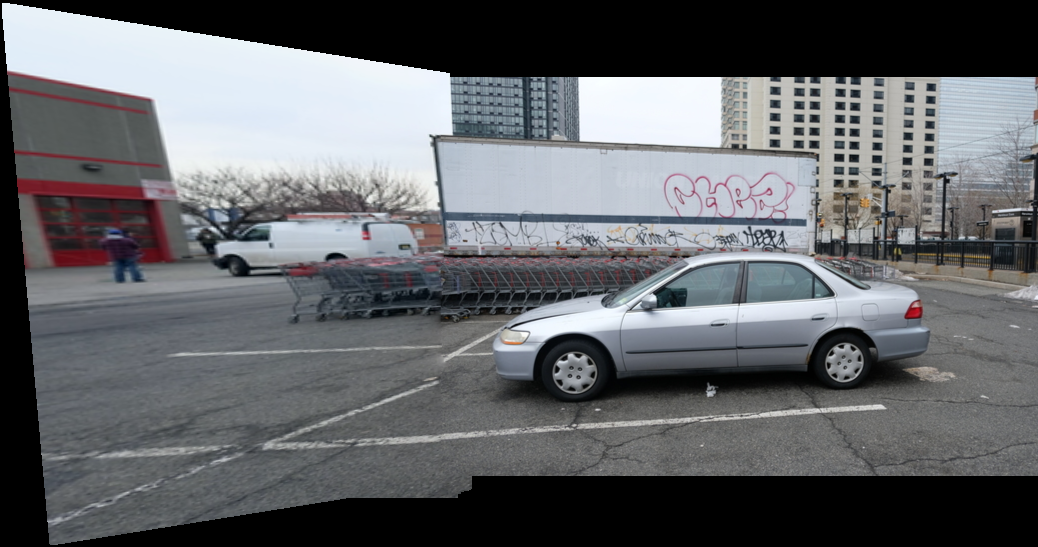}{``Parking lot'' dataset. Autostitch
			fails to stitch. APAP duplicates the car's hood , tears a background building, and introduces a corner in the roof of the trailer. Photoshop duplicates the front half of the
			car. NIS has substantial ghosting. Our result cuts out a part of a parking
			stall line, but avoids duplicating the car.  }
		
		\bigresultsfigure{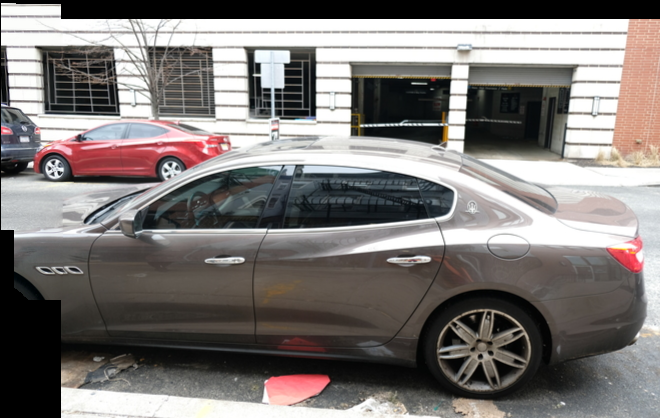}{``Cars'' dataset.
			Autostitch fails to stitch.
			APAP and Photoshop shorten the car. APAP also introduces substantial curvature into the background building. NIS has substantial ghosting and shortens the car. Our result deletes part of the hood and front wheel; however, it is the only result which produces an artifact-free car body.
		}
		
		\begin{centering}
			\begin{subfigure}{0.5\textwidth}
				\includegraphics[width=0.45\textwidth]{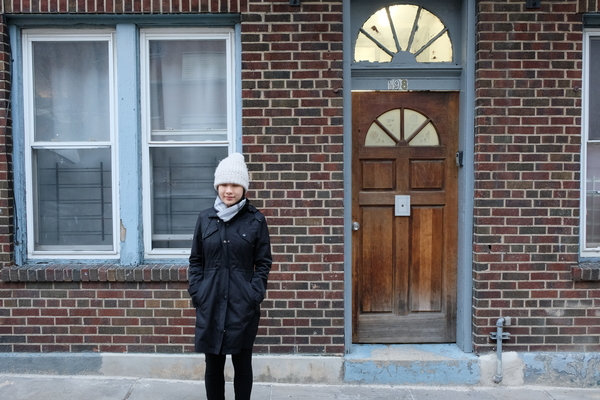}
				\includegraphics[width=0.45\textwidth]{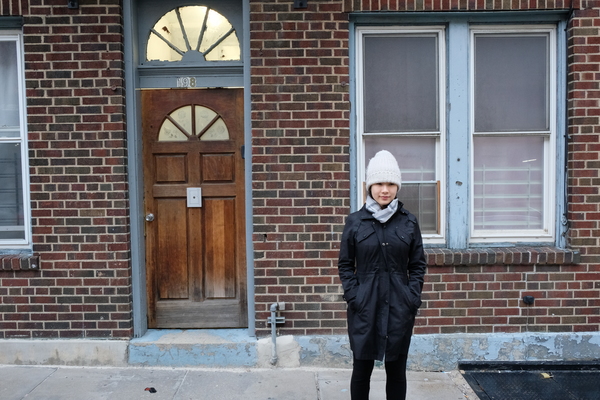}
				\caption{Input images}
			\end{subfigure}
			\begin{subfigure}{0.35\textwidth}
				\includegraphics[width=\textwidth]{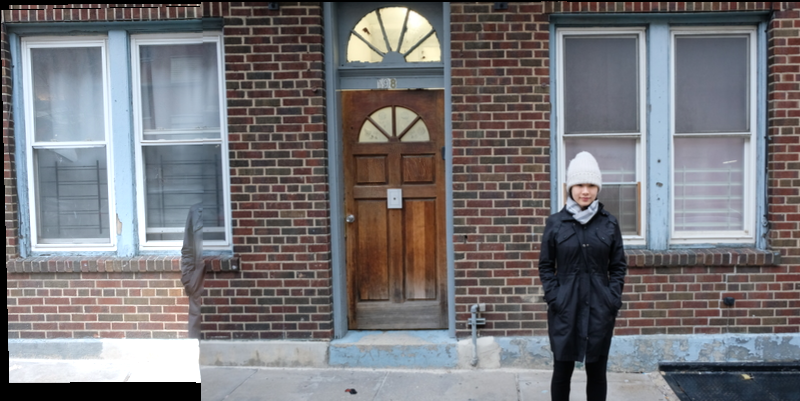}
				\caption{Our result}
			\end{subfigure}
			\caption{\label{fig:tearing_example} An example of tearing and duplication produced by our method.}
		\end{centering}
	\end{figure}%
	
	\noindent \textbf{Quantitative evaluation.} The only quantitative metric used by previous stitching papers is seam quality (MRF energy). However, as
	we have shown, local seam quality is not indicative of stitch quality. Also,
	this technique requires the user to know the seam location, which precludes it
	from being run on black-box algorithms like Photoshop. Here we attempt to define
	a metric to address these problems.
	
	We first observe that stitching can be viewed as a form of view synthesis with weaker assumptions regarding the camera placement or type. With this connection in mind, we redefine perspective stitching as extending the field of view of a reference image using the information in the candidate images. This redefintion naturally leads to an evaluation technique. We crop part of the reference image and then stitch the cropped image with the candidate image. This cropped region serves as a ground truth, which we can compare against the appropriate location in the stitch result. Note that in perspective stitching, the reference image's size is not altered so we know the exact area where the cropped region should be. We then calculate MS-SSIM \cite{MSSIM} or PSNR.
	
	We report this evaluation for 2 examples in Table~\ref{tab:evaluation}: 50 pixels are cropped off the edge of the reference images in Stop Sign (left side of first image for Figure~\ref{fig:stop_sign_1}) and Graffiti Building (right side of first image for Figure~\ref{fig:graffiti_building_1}). The stitch results for the cropped images appear almost identical to the stitch results for the whole images. Best score shown inbold. ``Ground Truth'' compares only the ground truth region to the
	appropriate location, while ``Uncropped Reference'' compares the uncropped reference.
	
	\begin{table}[t]
		\begin{center}
			\caption{Evaluation scores for different algorithms.}
			\label{tab:evaluation}
			\begin{tabular}{| l | l | l || c | c | c |}
				\hline
				Image                & Comparison          & Metric  & Ours & APAP~\cite{zaragoza2013projective} & Photoshop~\cite{Adobe:Photomerge} \\\hline
				Stop Sign           & Ground Truth Region & MS-SSIM & 0.6851 & 0.6573 & \bf{0.6861} \\ 
				&                     & PSNR    & \bf{19.4943} & 17.7073 & 18.9996 \\
				& Uncropped Reference & MS-SSIM & \bf{0.9354} & 0.8981 & 0.9108 \\ 
				&                     & PSNR    & \bf{23.0006} & 20.3533 & 20.9238 \\
				\hline
				Graffiti Building   & Ground Truth Region & MS-SSIM & \bf{0.4636} & 0.3747 & 0.1250 \\ 
				&                     & PSNR    & \bf{14.9983} & 13.1269 & 9.6520 \\
				& Uncropped Reference & MS-SSIM & \bf{0.9253} & 0.5737 & 0.8541 \\ 
				&                     & PSNR    & \bf{24.7637} & 14.8298 & 18.7102 \\ \hline
			\end{tabular}
		\end{center}
	\end{table}
	
	
	Note that for Stop Sign, all algorithms performed reasonably in Ground Truth Region. However, both APAP and Photoshop include a duplicate
	of the stop sign that lowers their values for Uncropped Reference.

	\section{Conclusions, limitations, and future work}
	\label{sec:extensions}
	
	We have demonstrated a novel formulation of the image stitching problem in which
	multiple candidate registrations are used. We have generalized
	MRF seam finding to this setting and proposed new terms to combat common artifacts
	such as object duplication. Our techniques outperform existing
	algorithms in large parallax and motion scenarios. 
	
	Our methods naturally generalize to other stitching surfaces such as
	cylinders or spheres via modifications to the warping function. 
	Three or more input images can be handled by proposing multiple registrations 
	of each candidate image, and letting the seam finder composite them.
	A potential problem is the presence of undetected sparse correspondences, which can
	lead to duplications or tears (Figure~\ref{fig:tearing_example}). The use of dense
	correspondences may remedy this issue,
	but our preliminary experiments suggest that optical flows cannot easily capture motion
	in input images with large disparities, and do not produce correspondences of sufficient
	quality. A second issue is that it is unclear whether to populate regions of the output
	mosaic when only data from a single candidate image is present, as the constrained
	choice of candidate here may conflict with choices made in other regions of the mosaic.
	This can to some extent be handled with modifications to the data term, but compared to
	traditional methods, scene content may be lost. One example of this occurs in
	Figure~\ref{fig:cars_1}, where the front wheel of the car is omitted in the final
	output. These problems remain exciting challenges for future work.

	\subsubsection{Acknowledgements} This research was supported by NSF grants IIS-1161860 and IIS-1447473 and by a Google Faculty Research Award. We thank Connie Choi for help collecting images.
	
	\clearpage
	\newpage
	{\small
		\bibliographystyle{splncs}
		\bibliography{ourbib}

\begin{thebibliography}{10}

\bibitem{Lin2016}
Lin, K., Jiang, N., Cheong, L.F., Do, M., Lu, J.:
\newblock Seagull: Seam-guided local alignment for parallax-tolerant image
  stitching.
\newblock In: ECCV. (2016)  370--385

\bibitem{Zhang_2014_CVPR}
Zhang, F., Liu, F.:
\newblock Parallax-tolerant image stitching.
\newblock In: CVPR. (2014)  3262--3269

\bibitem{SergeWhatWentWhere}
Willis, J., Argawal, S., Serge, B.:
\newblock What went where.
\newblock In: CVPR. (2003)  37--44

\bibitem{WangAdelson}
Wang, J., Adelson, E.:
\newblock Representing moving images with layers.
\newblock TIP \textbf{3}(5) (1994)  625--638

\bibitem{Kwatra:2003}
Kwatra, V., Sch\"{o}dl, A., Essa, I., Turk, G., Bobick, A.:
\newblock Graphcut textures: Image and video synthesis using graph cuts.
\newblock SIGGRAPH \textbf{22}(3) (2003)  277--286

\bibitem{Szeliski:tutorial:2006}
Szeliski, R.:
\newblock Image alignment and stitching: A tutorial.
\newblock Foundations and Trends in Computer Graphics and Vision \textbf{2}(1)
  (2007)  1--104

\bibitem{Szeliski10}
Szeliski, R.:
\newblock Computer Vision: Algorithms and Applications.
\newblock Springer (2010)

\bibitem{Brown_IJCV_2007}
Brown, M., Lowe, D.G.:
\newblock Automatic panoramic image stitching using invariant features.
\newblock IJCV \textbf{74}(1) (2007)  59--73

\bibitem{Perez:Poisson2003}
Perez, P., Gangnet, M., Blake, A.:
\newblock Poisson image editing.
\newblock SIGGRAPH (2003)  313--318

\bibitem{Chen2016}
Chen, Y.S., Chuang, Y.Y.:
\newblock Natural image stitching with the global similarity prior.
\newblock In: ECCV. (2016)  186--201

\bibitem{Adobe:Photomerge}
Adobe:
\newblock Create panoramic images with photomerge.
\newblock
  \url{https://helpx.adobe.com/in/photoshop/using/create-panoramic-images-photomerge.html}.
  Accessed: 2018-07-25.

\bibitem{AdobeHelp:2018}
Adobe:
\newblock Create and edit panoramic images.
\newblock
  \url{https://helpx.adobe.com/photoshop/using/create-panoramic-images-photomerge.html}.
  Accessed: 2018-07-25.

\bibitem{shum2001construction}
Shum, H.Y., Szeliski, R.:
\newblock Construction of panoramic image mosaics with global and local
  alignment.
\newblock In: Panoramic vision. (2001)  227--268

\bibitem{zaragoza2013projective}
Zaragoza, J., Chin, T.J., Brown, M.S., Suter, D.:
\newblock As-projective-as-possible image stitching with moving dlt.
\newblock In: CVPR. (2013)  2339--2346

\bibitem{Lin_2015_CVPR}
Lin, C.C., Pankanti, S.U., Natesan~Ramamurthy, K., Aravkin, A.Y.:
\newblock Adaptive as-natural-as-possible image stitching.
\newblock In: CVPR. (2015)  1155--1163

\bibitem{SZSVKATR:PAMI08}
Szeliski, R., Zabih, R., Scharstein, D., Veksler, O., Kolmogorov, V., Agarwala,
  A., Tappen, M., Rother, C.:
\newblock A comparative study of energy minimization methods for {Markov}
  {Random} {Fields}.
\newblock TPAMI \textbf{30}(6) (2008)  1068--1080

\bibitem{Agarwala:2004:IDP:1015706.1015718}
Agarwala, A., Dontcheva, M., Agrawala, M., Drucker, S., Colburn, A., Curless,
  B., Salesin, D., Cohen, M.:
\newblock Interactive digital photomontage.
\newblock SIGGRAPH \textbf{23}(3) (2004)  294--302

\bibitem{Burt_TOG_83}
Burt, P., Adelson, E.:
\newblock A multiresolution spline with application to image mosaics.
\newblock SIGGRAPH \textbf{2}(4) (1983)  217--236

\bibitem{opencv_library}
Bradski, G.:
\newblock {The OpenCV Library}.
\newblock Dr. Dobb's Journal of Software Tools (2000)

\bibitem{ceres-solver}
Agarwal, S., Mierle, K., Others:
\newblock Ceres solver.
\newblock \url{http://ceres-solver.org}. Accessed: 2018-07-25.

\bibitem{Liu:SIGGRAPH09}
Liu, F., Gleicher, M., Jin, H., Agarwala, A.:
\newblock Content-preserving warps for {3D} video stabilization.
\newblock SIGGRAPH \textbf{28}(3) (2009)

\bibitem{BVZ:PAMI01}
Boykov, Y., Veksler, O., Zabih, R.:
\newblock Fast approximate energy minimization via graph cuts.
\newblock TPAMI \textbf{23}(11) (2001)  1222--1239

\bibitem{Weinzaepfel_2013_ICCV}
Weinzaepfel, P., Revaud, J., Harchaoui, Z., Schmid, C.:
\newblock Deepflow: Large displacement optical flow with deep matching.
\newblock In: ICCV. (2013)  1385--1392

\bibitem{MSSIM}
Wang, Z., Simoncelli, E., Bovik, A.:
\newblock Multiscale structural similarity for image quality assessment.
\newblock In: Asilomar Conference on Signals, Systems and Computers. (2004)
  1398--1402

\end{thebibliography}
	}
	
\end{document}